%% file: _main.tex
\documentclass[sigconf,screen]{acmart}

\usepackage{hyperref}
\usepackage{enumitem}
\usepackage{nicefrac}
\usepackage{multirow}
\usepackage{color}
\usepackage{import}
\usepackage{bm}
\usepackage{subfigure}
\usepackage{adjustbox}
\usepackage{array}
\usepackage{balance}
\newcommand{\oR}{\textsuperscript{\textregistered}\ }

\AtBeginDocument{}

\copyrightyear{2026}
\acmYear{2026}
\setcopyright{cc}
\acmConference[FPGA '26]{Proceedings of the 2026 ACM/SIGDA International Symposium on Field Programmable Gate Arrays}{February 22--24, 2026}{Seaside, CA, USA}
\acmBooktitle{Proceedings of the 2026 ACM/SIGDA International Symposium on Field Programmable Gate Arrays (FPGA '26), February 22--24, 2026, Seaside, CA, USA}
\acmPrice{}
\acmDOI{10.1145/3748173.3779200}
\acmISBN{979-8-4007-2079-6/2026/02}

\begin{document}

\title{HGQ: High Granularity Quantization for Real-time Neural Networks on FPGAs}

\author{Chang Sun}
\affiliation{
    \institution{California Institute of Technology}
    \city{Pasadena}
    \state{California}
    \country{USA}
}
\email{chsun@cern.ch}
\orcid{0000-0003-2774-175X}
\authornote{Corresponding author; Partial work done while at ETH Zurich.}

\author{Zhiqiang Que}
\affiliation{
    \institution{Imperial College London}
    \city{London}
    \country{UK}
}
\email{z.que@imperial.ac.uk}
\orcid{0000-0002-9263-6529}

\author{Thea \AA rrestad}
\affiliation{
    \institution{ETH Zurich}
    \city{Zurich}
    \state{Zurich}
    \country{Switzerland}
}
\email{thea.aarrestad@cern.ch}
\orcid{0000-0002-7671-243X}

\author{Vladimir Loncar}
\affiliation{
    \institution{Institute of Physics Belgrade}
    \city{Belgrade}
    \country{Serbia}
}
\additionalaffiliation{
    \institution{CERN}
    \city{Geneva}
    \country{Switzerland}
}
\email{vloncar@ipb.ac.rs}
\orcid{0000-0003-3651-0232}

\author{Jennifer Ngadiuba}
\affiliation{
    \institution{Fermilab}
    \city{Batavia}
    \state{Illinois}
    \country{USA}
}
\orcid{0000-0002-0055-2935}
\email{jennifer.ngadiuba@cern.ch}

\author{Wayne Luk}
\affiliation{
    \institution{Imperial College London}
    \city{London}
    \country{UK}
}
\email{w.luk@imperial.ac.uk}
\orcid{0000-0002-6750-927X}

\author{Maria Spiropulu}
\affiliation{
    \institution{California Institute of Technology}
    \city{Pasadena}
    \state{California}
    \country{USA}
}
\email{smaria@caltech.edu}
\orcid{0000-0001-8172-7081}

\renewcommand{\shortauthors}{Chang Sun et al.}
\newcommand{\fmax}{$\mathrm{F}_\mathrm{max}$}

\begin{abstract}
    Neural networks with sub-microsecond inference latency are required by many critical applications.
    Targeting such applications deployed on FPGAs, we present High Granularity Quantization (HGQ), a quantization-aware training framework that optimizes parameter bit-widths through gradient descent.
    Unlike conventional methods, HGQ determines the optimal bit-width for each parameter independently, making it suitable for hardware platforms supporting heterogeneous arbitrary precision arithmetic.
    In our experiments, HGQ shows superior performance compared to existing network compression methods, achieving orders of magnitude reduction in resource consumption and latency while maintaining the accuracy on several benchmark tasks.
    These improvements enable the deployment of complex models previously infeasible due to resource or latency constraints.
    HGQ is open-source\footnote{https://github.com/calad0i/hgq2} and is used for developing next-generation trigger systems at the CERN ATLAS and CMS experiments for particle physics, enabling the use of advanced machine learning models for real-time data selection with sub-microsecond latency.
\end{abstract}

\ccsdesc[500]{Hardware~Hardware-software codesign}
\ccsdesc[500]{Computing methodologies~Neural networks}
\ccsdesc[300]{Applied computing~Physics}

\keywords{Quantization-aware training, FPGA, Real-time inference, Neural networks, Hardware-software codesign, Low-latency, Quantization}

\maketitle

\section{Introduction}
\label{sec:introduction}

Edge computing has significantly increased the importance of real-time Deep Neural Network (DNN) inference on specialized hardware~\cite{edge-survey}.
While the typical latency threshold for real-time inference applications is $\mathcal{O}(1)$~ms~\cite{realtime1,realtime2,realtime3}, some critical applications need few-microsecond to sub-microsecond inference latency with high throughput requirements.
The most prominent example of such applications is the level-1 \textit{trigger}, a real-time data filtering system at the CERN Large Hadron Collider (LHC)~\cite{lhc1995large} experiment, which requires end-to-end decision latencies not exceeding a few microseconds~\cite{cms-tdr-021,atlas-tdr-029}.
Similar latency and resource constraints also apply to other accelerator experiments, such as Belle II~\cite{belle2-1,belle2-2}, where a proposed neural network-based trigger of $\sim 300$ns latency is required~\cite{belle2-tk,belle2-nnt}.

Similar requirements are also found in other scientific domains, such as neutrino and dark matter searches~\cite{fastml-for-science}, quantum circuit control~\cite{qick}, gravitational wave detection~\cite{que2021accelerating} and magnetic confinement fusion control~\cite{jet-tokamak}.
Outside the scientific computing domain, real-time inference with (sub-)microsecond latency is needed in many applications, such as event cameras~\cite{evtcam-sub0, evtcam-sub1, evtcam-sub2,evtcam-denoise, evtcam-review}, radar signal processing~\cite{radar}, augmented/virtual reality applications~\cite{motion-to-photons}, and high-frequency trading~\cite{hft-ethernet, hft-speed}.
However, it is extremely challenging to achieve (sub-)microsecond end-to-end latency for DNN inference on traditional hardware.
Accuracy typically benefits from larger, deeper models, whereas ultra-low latency requires highly parallel, even fully unrolled, implementation on specialized hardware, like FPGAs.
While using FPGAs mitigates such latency requirements, it can lead to high resource consumption, creating a tension between model size and hardware cost.
This tension is often acute in scientific and control applications, e.g., for L1 triggers at colliders and fast control loops, where both latency and throughput constraints must be met simultaneously under hard resource caps.

To address such challenges, recent work explores several directions. First, fully unrolled pipelines can deliver large latency reductions by spatially mapping every operation, but they scale poorly in resources as models grow~\cite{nn-fpga, tridgell2019unrolling}.
In the same spirit of pushing logic-level parallelism, LUT-based DNNs map neurons or subfunctions to learned logical look-up operations which are further mapped to physical LUT primitives, yielding DSP-free, deeply pipelined sub-$\mu$s designs on compact models.
However, existing efforts mostly target Multilayer Perceptron (MLP) networks at small scales, typically at most a few thousand learned LUTs, confine computation to LUTs which underuses other on-chip resources such as DSPs and increases routing pressure as models scale, and rely on largely AMD/Xilinx-centric toolflows~\cite{lutnet, polylut, neuralut, treelut, reducedlut,amigolut}, leaving scalability, architectural breadth, and cross-vendor portability as open challenges.

Quantization-aware training (QAT) is often used to improve hardware efficiency for neural networks, and frameworks such as QKeras~\cite{qkeras} are widely used for training FPGA-targeted DNNs with fixed-point quantization.
However, the compression ratio is limited with QAT when the model performance degradation is required to be small, and the choice of bit-widths itself becomes an important hyperparameter that needs to be optimized.
Differentiable quantization, such as DiffQ~\cite{diffq} has been proposed to optimize the bit-widths by parametrized noise injection, and LSQ~\cite{lsq} uses a straight-through estimator (STE) to optimize the quantization scale.
However, these methods still operate at coarse granularity, i.e., per-layer or per-channel, and do not incorporate FPGA-targeted resource models.
Moreover, these methods also come with their own challenges, such as extreme instability under low bit-widths~\cite{diffq}, or hardware-unfriendly quantization schemes~\cite{lsq}.
Consequently, there remains a gap for methods that (a)~learn bit-widths at high granularity and (b)~explicitly trade off task accuracy against on-chip cost for FPGA-targeted DNNs.

\begin{figure}
    \centering
    \includegraphics[width=0.8\linewidth]{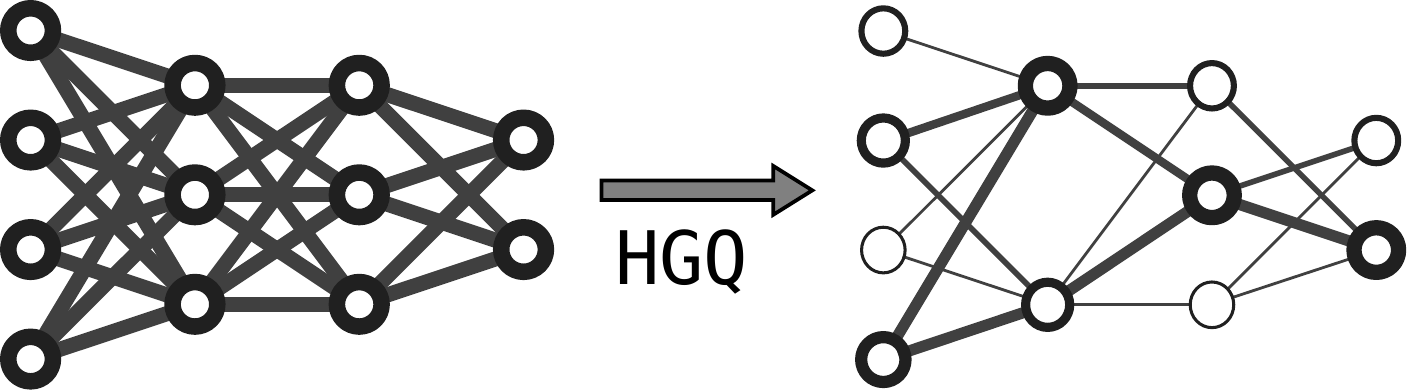}
    \caption{
        An illustration of the HGQ quantization scheme on weights and activations.
        In this example, each weight and activation has its own learnable bit-width.
        Parameters with zero bit-width are effectively pruned.
    }
    \label{fig:quant_illu}
    \vspace{-0.7cm}
    \Description{A figure showing two neural networks side by side. The left network shows weights and activations with uniform, high bit-widths, and the right network shows weights and activations with varying, sometimes pruned (bit-width zero) parameters. An arrow points from the left to the right network with HGQ labeled above it.}
\end{figure}

To address this gap, we propose High Granularity Quantization (HGQ), a QAT method tailored for FPGA-targeted neural networks.
HGQ has two key components: (i)~differentiable fixed-point quantization with learnable bit-widths optimized at an arbitrarily fine granularity (i.e., up to per-parameter, but can also be per channel/kernel/tile).
An intuitive illustration of the per-weight and per-activation quantization is shown in Fig.~\ref{fig:quant_illu}.
During training, we relax inherently discrete bit-widths to continuous surrogates, attach gradients to these surrogates, and round to integers in the forward pass before quantization so that standard gradient descent can allocate higher precision only where it most benefits.
(ii)~A differentiable on-chip resource usage estimator that acts as a regularizer during training.
This joint treatment encourages accuracy where it matters while penalizing resource-heavy configurations, producing FPGA-efficient quantizers that help meet (sub-)microsecond latency under tight hardware budgets without sacrificing competitive accuracy.

Unlike LUT-based logic mappings, HGQ preserves arithmetic structure, which enables the use of established neural network architectures and training methods.
The synthesis backend of HGQ can flexibly mix LUTs and DSP primitives for the implementation and is designed to scale to larger models than current LUT-based methods, making it a more versatile solution for microsecond DNN inference on FPGAs.
For very small networks that HGQ can produce a solution with a few thousand LUTs or less (such as some in Tab.~\ref{tab:hlf_jsc}), existing LUT-based approaches may achieve better absolute resource efficiency and latency in comparison to HGQ, while both approaches can meet realistic latency constraints in this regime.
However, for larger models with microsecond-level latency constraints that the smallest designs from HGQ require tens of thousands of LUTs or more (as shown in Tab.~\ref{tab:plf_jsc} and Tab.~\ref{tab:tgc_svhn}), the current LUT-based methods are not directly applicable due to scalability challenges, and HGQ provides the best resource-accuracy trade-offs among existing methods for such models.
Nevertheless, we note that while HGQ is intended for models of larger scales than the LUT-based ones, the target models are still compact compared to typical machine learning models, as the microsecond-order latency requirement still forces the model to fit reasonably on a single FPGA when fully-or-partially unrolled.

We developed a reusable HGQ framework and released it as a free and open-source software under the LGPL-3 license.
The Vivado/Vitis\oR FPGA back-ends are supported through integration with \texttt{da4ml}~\cite{da4ml} and \texttt{hls4ml}~\cite{hls4ml}, free and open-source libraries that transform machine learning algorithms into hardware designs.
In particular, \texttt{da4ml} targets designs with initiation interval (II) of 1, optimizes the design with distributed arithmetic (DA), and can emit Verilog and VHDL code for various back-ends, while \texttt{hls4ml} is more general and emits HLS code for multiple back-ends, including Vivado/Vitis\oR HLS and Intel\oR HLS.
The end-to-end workflow is illustrated in Fig.~\ref{fig:workflow}.

To the best of our knowledge, this paper is the first to propose differentiable, quantization-aware codesign that learns per-parameter bit-widths and jointly optimizes accuracy and FPGA resources for sub-microsecond DNNs.
The main contributions of this work are:
\begin{itemize}[noitemsep,topsep=5pt]
    \item HGQ, a QAT approach that achieves differentiable fixed-point quantization with learnable bit-widths optimized at arbitrary granularity for hardware-efficient DNNs on FPGAs. It automatically includes pruning by assigning zero bit-width to pruned parameters.
    \item Releasing HGQ as a free and open-source library, with an end-to-end software-hardware codesign workflow closely integrated with the public-domain \texttt{da4ml} and \texttt{hls4ml} tools.
    \item A comprehensive evaluation of the proposed framework. HGQ provides significant resource and latency improvement while maintaining model accuracy compared to other quantization methods when deploying on FPGAs.
\end{itemize}

\section{Background and Related Work}

Efforts in recent years have focused on algorithmic efficiency, using compact architecture, pruning and especially quantization~\cite{compression1, compression2}.
Quantization is widely adopted for compressing DNNs for specialized hardware such as FPGAs or ASICs with low precision quantization, such as binary or ternary, enhancing throughput and latency.
Key examples include DoReFa Net~\cite{dorefa}, ABC-net~\cite{abc-net}, Binaryconnect~\cite{binaryconnect}, XNOR-net~\cite{xnor-net}, TWN~\cite{twn}, TTQ~\cite{ttq}, and \cite{tridgell2019unrolling}.
With the same principle, several studies~\cite{lq-similar-1, lq-similar-2, lq-net, mbn-loss,mix-match} utilize multi-bit network designs that represent weights through binary bases and floating-point values.

Heterogeneous quantization assigns layer/channel-specific precision to reduce accuracy loss. In particular, \cite{haq} and \cite{autoq} use reinforcement learning to find optimal bit-widths.
\cite{hawq, hawq-v2, pyhessian, q-bert, obq} focus on optimizing bit-widths with second-order approximations of the loss function.
DNAS~\cite{dnas} and AutoQKeras~\cite{qkeras} optimize bit-widths and network architecture simultaneously with stochastic sampling, or gradient-free search.
Similarly, MetaML~\cite{metaml, metamlpro} applies iterative optimizations to various hyperparameters. FILM-QNN~\cite{film} targets hardware-friendly bit-widths for CNNs.
Heterogeneous quantization at sub-layer/channel granularity has also been studied. RVQuant~\cite{rvquant}, BitsandBytes~\cite{bnb}, SpQR~\cite{spqr}, and SqueezeLLM~\cite{squeeze_llm} keep a small set of outlier weights at higher precision.
These approaches primarily target weight size reduction of larger models, rather than efficient inference on specialized hardware.

On the subfield of real-time inference of neural networks on FPGAs, the state-of-the-art models in the LHC community are produced with the QKeras-hls4ml workflow~\cite{qkeras,hls4ml, snac, ds-fpga}.
Although many optimizations, such as neural architecture search, pruning strategies or hybrid training schedules are proposed, the core QAT still adopts a conventional fixed-precision, uniform quantization scheme. This is primarily due to constraints from the FPGA platforms together with latency requirements, which effectively forbids the use of non-uniform quantization and grouped scaling factors.

Closely related to this work, \texttt{QKeras}~\cite{qkeras}, built on \texttt{Keras} with \texttt{hls4ml}~\cite{hls4ml} for deployment, trains networks with hardware-friendly fixed-point weights and activations. Its \texttt{AutoQKeras} feature tunes layer-wise quantization via gradient-free search.
QKeras is the \textit{de facto} framework for training neural networks for sub-microsecond DNN inference on FPGAs in the LHC community, and it is widely used in the CERN LHC experiments~\cite{qkeras, hls4ml, snac, ds-fpga,tgc, axol1tl, cicada}.
Brevitas~\cite{brevitas}, the PyTorch~\cite{torch} counterpart, is commonly used with AMD's \texttt{FINN} or \texttt{FINN-R}~\cite{FINN, FINNR} flow for AMD\oR FPGAs.

LUT-based DNN inference~\cite{lutnet, polylut, neuralut, reducedlut,amigolut,neuralut-assemble} instead maps computations directly to FPGA logic look-up tables.

Some approaches begin from standard NNs (e.g., NeuraLUT~\cite{neuralut}), while more advanced ones (e.g., DWN~\cite{dwn}, NeuraLUT-Assemble~\cite{neuralut-assemble}) train or finetune directly in the LUT domain.
These methods deliver extremely low latency and high throughput on compact models, at the cost of moderate accuracy degradation versus floating point.
Moreover, to the best of our knowledge, these methods have only been demonstrated on small models of a few thousand LUTs, and it is unclear if they can be applied to larger models of order of $\sim100$k LUTs, which are shown to be necessary for many real-world applications~\cite{llgnn,ds-fpga,snac}.
When scaling to larger models, the sparse gradient in the look-up table space may lead to further challenges in training stability and convergence.
Our method is orthogonal to these LUT-based approaches, offering an alternative path to sub-microsecond FPGA inference.
Since several LUT-based methods also rely on quantization-aware training of the logic lookup tables, combining them with HGQ to further improve resource efficiency is a promising direction for future work.

On the FPGA deployment side, several open-source libraries exist for deploying machine learning models on FPGAs.
Among them, \texttt{hls4ml} is the most widely used open-source library for deploying machine learning models on FPGAs with sub-microsecond latency~\cite{hls4ml}.
It supports various neural network architectures, including fully-connected networks, convolutional networks, and some graph-based networks.
The framework has been adopted in production for the L1 trigger systems at the CMS experiment~\cite{axol1tl,cicada}.
\texttt{da4ml}~\cite{da4ml} is another open-source library for mapping a quantized neural network to FPGA designs, focusing on optimizing the constant-matrix-vector multiplications (CMVMs) in the network with DA~\cite{da}.
Instead of leaving the multiplication operations to be inferred by the HLS/synthesis tool, it explicitly transforms the CMVMs into optimized adder graphs with common subexpression elimination.
The framework can emit standard Verilog or VHDL code, or may be used as a plugin to \texttt{hls4ml} for  CMVM operations in various layers.
It is also used in production at the CMS L1 trigger system~\cite{axol1tl} in conjunction with \texttt{hls4ml}.
QONNX~\cite{qonnx} is a quantization extension to the ONNX~\cite{onnx} format, which provides a unified format for representing quantized neural networks, and acts as the common interface for \texttt{FINN}~\cite{FINN, FINNR} and \texttt{hls4ml}.
HGQ currently has partial support to export its optimized models to QONNX.
HGQ integrates seamlessly with both \texttt{hls4ml} and \texttt{da4ml}, acting as a hardware-friendly frontend that, given an accuracy target and resource budget, automatically explores mixed-precision configurations.

\section{High Granularity Quantization (HGQ)}

This section introduces the HGQ algorithm, which  consists of a differentiable, fixed-point quantization scheme promoting loss-awareness for lower bit-widths, and a differentiable on-chip resource estimator promoting lower resource consumption.

\subsection{Differentiable quantization}
\subsubsection{Fixed-point quantization scheme}
We adopt hardware-efficient fixed point representations.
In particular, we use a scheme compatible with the \texttt{ac/ap\_fixed} number schemes in \texttt{hlslib/vitis}.
We define the map $\mathrm{f}: \mathbb{R}\rightarrow\mathbb{Q}\subset\mathbb{R}$ as the quantizer, which takes five parameters -- $s$: (boolean) signed, $i$: (integer) integer bits, excluding sign, $f$: (integer) fractional bits, $r\_mode$: (choice) rounding mode, and $o\_mode$: (choice) overflow mode.
The representable values of a quantized number $x^q$ are $\mathbb{Q} = \left[-s \cdot 2^i, 2^i-2^{-f}\right]$ with a step size of $2^{-f}$.
The total width of the number is $w = s + i + f$.
The quantization operation is given by eq.~\eqref{eq:quant_op}, where $[*]$ is the rounding operation depending on $r\_mode$, while the overflow behavior depends on $o\_mode$.
Depending on $r\_mode$ and $o\_mode$, the quantization operation may incur different hardware overheads when performed on FPGAs.
Since the weights are known after training, the overhead is only a concern for the activations.
\begin{equation}
    \mathrm{q}(x) = \begin{cases}
        \left[x\cdot2^{f}\right]\cdot2^{-f}, & \text{if } x\in[-s\cdot 2^i, 2^i-2^{-f}] \\
        \mathrm{overflow}                    & \text{otherwise}
    \end{cases}
    \label{eq:quant_op}
\end{equation}

Throughout the network, we primarily use  $r\_mode=$\texttt{RND}, which recovers the round-to-nearest, ties up scheme ($\mathrm{round}(x) = \mathrm{floor}(x + 0.5)$) due to high stability and low overhead.
While \texttt{TRN} has the absolute minimal overhead which  truncates beyond the LSBs, it introduces an undesirable systematic bias into the quantization error.
While the additional addition required by \texttt{RND} is an overhead, it can frequently be fused into the bias addition operation in neural networks, thus the overhead introduced is negligible.

Both $o\_mode=$\texttt{SAT} (clipping) and $o\_mode=$\texttt{WRAP} (modulo) are supported in our framework.
The \texttt{WRAP} mode has minimal hardware overhead by truncating beyond the MSB, but this effectively performs a modulo operation, which is chaotic during training due to the discontinuity and the periodicity.
To mitigate this issue, we track only $f$ during training for values with $o\_mode=$\texttt{WRAP}, and determine $i$ before deployment.
For weights, this is trivial, and for activations, this is done by profiling the activation ranges on a representative dataset and statistically avoiding the overflows.
In practice, we find using the training dataset for this purpose would usually suffice.
In contrast, \texttt{SAT} mode is more stable during training and produces lower bit-widths in general for activations;
however, it is also associated with a high overhead, as it requires additional comparators and multiplexers to perform the clipping.
Including the overflow behavior, during training, the quantization operation is given by Eq.~\eqref{eq:train_quant}.
\begin{equation}
    \mathrm{q}_\mathrm{train}(x) = \begin{cases}
        \left[x\cdot2^{f}\right]\cdot2^{-f}                                                  & \text{if \texttt{WRAP}} \\
        \mathrm{clip}\left(\left[x\cdot2^{f}\right]\cdot2^{-f},-s\cdot 2^i,2^i-2^{-f}\right) & \text{if \texttt{SAT}}
    \end{cases}
    \label{eq:train_quant}
\end{equation}

The rounding operation $[*]$ is non-differentiable.
During training, we use the Straight Through Estimator (STE)~\cite{ste} to approximate the gradients of the quantization operation to tune the parameters of the network.
The STE is a technique commonly used in training quantized neural networks~\cite{qkeras,a2qp}, as it provides a simple and effective way to estimate gradients.
By applying STE to the rounding operation, we approximate its gradient as identity, i.e., ${\partial \mathrm{q}(x)}/{\partial x} = 1$.
This allows gradients to flow through the quantization operation during backpropagation, allowing optimization of the network parameters.

\subsubsection{Connection to Pruning}

Eq.~\eqref{eq:train_quant} shows that the quantized value is zero if $|x|<2^{-f-1}$.
As $f\in\mathbb{Z}$, a sufficiently small $f$ will cause the corresponding parameters in the network to vanish, effectively pruning these parameters.
Thus, HGQ automates network pruning during training by assigning a distinct bit-width to each parameter group in the network, taking into account both model performance and resource consumption.
When the granularity for quantization is on a per-parameter basis, a fully unstructured pruning is automatically performed.

\subsection{Surrogate gradients}

To optimize bit-widths with per-parameter granularity, we need to make the bit-widths differentiable with gradients attached to them.
While bit-widths are inherently discrete, we relax them to be continuous values and round them to integers before the quantization operations.

The gradients from the lower and upper bounds for the \texttt{SAT} overflow mode are straightforward, while the gradient from the $2^{-f}\cdot [2^f\cdot x]$ term in both \texttt{SAT} and \texttt{WRAP} modes requires careful treatment.

To obtain the surrogate gradient from the $2^{-f}\cdot [2^f\cdot x]$ term, we first consider a parameter $x$ (e.g., weight or activation) in the network and its corresponding fractional bit-width $f$.
The associated quantization error $\delta_f$ related to this term can be expressed as $\delta_f\equiv x-\mathrm{f}^q(x)=x-\left[x\cdot2^{f}\right]\cdot2^{-f}$.
During training, let $x$  be a random variable following a smooth distribution $\mathbb{D}_x$, such that the variance of $\mathbb{D}_x$ is significantly larger than the quantization error $\delta_f$ so  that the quantization error follows a uniform distribution:
\begin{equation}
    \delta_f\sim\mathrm{Uniform}(-2^{-f-1},2^{-f-1}).
    \label{eq:quant_error}
\end{equation}
Let the loss of the network be $\mathcal{L}$. The gradient of $f$ with respect to $\mathcal{L}$ is given by
\begin{equation}
    \frac{\partial\mathcal{L}}{\partial f} = \frac{\partial\mathcal{L}}{\partial\delta_f}\cdot\frac{\partial\delta_f}{\partial f}.
    \label{eq:grad_obj_0}
\end{equation}

In this expression, the first term $\frac{\partial\mathcal{L}}{\partial\delta_f}$ can be obtained from backpropagation.
The second term $\frac{\partial\delta_f}{\partial f}$ is not well-defined, since $f$ can only take integer values for a properly defined quantizer and thus has no gradient.
To address this issue, we propose a surrogate gradient method that assigns a gradient to $f$ only on integer values.
We can now express the loss as a function of the weights $\bm\theta$ and all the quantization errors $\bm\delta$, $\mathcal{L}(\bm\theta,\bm\delta)$.
We assume that the loss function is sensitive to the magnitude of the quantization errors, but not their signs, i.e. $\mathcal{L}(\bm\theta,|\bm\delta|)$ with $|\bm\delta|$ being the element-wise absolute value of $\bm\delta$.

For a parameter $x\sim\mathcal{D}_x$ to be quantized with $f\in\mathbb{Z}$ fractional bits, the corresponding absolute quantization error is $|\delta_f|\equiv |x-\mathrm{f}^q_f(x)|\sim\mathrm{Uniform}(0,2^{-f-1})$.
By increasing $f$ by one, we obtain the absolute quantization error $|\delta_{f+1}|$ as a function of $f$ and $|\delta_f|$:
\begin{equation}
    |\delta_{f+1}| =
    \begin{cases}
        |\delta_f|            & |\delta_f| \le 2^{-f-2} \\
        2^{-f-1} - |\delta_f| & |\delta_f| > 2^{-f-2}   \\
    \end{cases}.
    \label{eq:delta_f}
\end{equation}

A straightforward way to obtain the gradient of $|\delta_f|$ with respect to $f$ is to use the finite difference approximation
\begin{equation}
    \frac{\partial |\delta_f|}{\partial f} \leftarrow |\delta_{f+1}| - |\delta_f|.
    \label{eq:finite_diff}
\end{equation}

However, as the absolute quantization error is bounded by a geometric sequence of $2^{-f-1}$, using a linear difference for approximation may be suboptimal.
Instead, we use the following heuristic expression to approximate the gradient, which recovers Eq.~\eqref{eq:finite_diff} in the limit of $|\delta_{f+1}|-|\delta_f|\rightarrow0$:
\begin{equation}
    \frac{\partial |\delta_f|}{\partial f} \leftarrow \log\frac{|\delta_{f+1}|}{|\delta_f|}\cdot|\delta_f|.
    \label{eq:finite_diff2}
\end{equation}

Expressing the ratio of $|\delta_{f+1}|$ and $|\delta_f|$ as a function of $|\delta_f|$, we have
\begin{equation}
    \frac{|\delta_{f+1}|}{|\delta_f|} =
    \begin{cases}
        1                             & |\delta_f| \le 2^{-f-2} \\
        \frac{2^{-f-1}}{|\delta_f|}-1 & |\delta_f| > 2^{-f-2}   \\
    \end{cases}.
    \label{eq:ratio}
\end{equation}

While one may get a surrogate gradient by combining Eq.~\eqref{eq:finite_diff2} and Eq.~\eqref{eq:ratio}, using the local relations as expressed in Eq.~\eqref{eq:ratio} between $|\delta_{f+1}|$ and $|\delta_f|$ would lead to a loss (gradient) landscape for $f$ with extensive high-frequency components;
such landscape is difficult to optimize. To mitigate this issue, we smooth out the loss (gradient) landscape by taking the expectation of the first term of Eq.~\eqref{eq:finite_diff2} over $|\delta_f|\sim\mathrm{Uniform}(0,2^{-f-1})$:
\begin{equation}
    \mathbb{E}_{|\delta_f|}\left[\log\frac{|\delta_{f+1}|}{|\delta_f|}\right] =-\log2.
    \label{eq:minus_log2}
\end{equation}

By substituting Eq.~\eqref{eq:minus_log2} into Eq.~\eqref{eq:finite_diff2}, and adding a $\mathrm{sign}(\delta_f)$ term on both sides, we have
\begin{equation}
    \frac{\partial\delta_f}{\partial f} \leftarrow -\log2\cdot\delta_f.
\end{equation}

The resulting gradient is smooth and easy to optimize.
Intuitively, it is the increment of the loss due to quantization with $f$ fractional bits, scaled by a factor of $\log2$.

\begin{figure*}[htbp]
    \centering
    \includegraphics[width=0.8\textwidth]{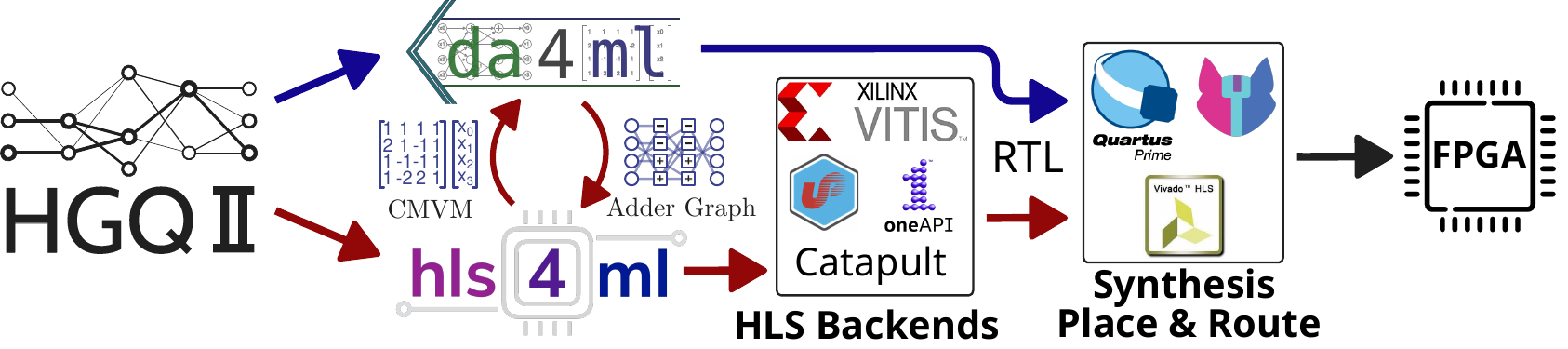}
    \vspace{-0.1cm}
    \caption{
        Overall workflow of the HGQ framework.
        The dark blue flow uses the \texttt{da4ml} backend, and the brown flow uses the \texttt{hls4ml} backend with optional DA optimizations.
    }
    \label{fig:workflow}
    \vspace{-0.3cm}
    \Description{A flowchart showing the two workflows of the HGQ-two framework. In the RTL generation flow with da4ml backend, the model is passed from HGQ-two to da4ml, the RTL code generated from da4ml is passed to synthesis and place and route backends to generate the FPGA bitstream. In the HLS generation flow with hls4ml, hls4ml may optionally send the CMVM operations to da4ml for optimization and receive the generated adder graph. The HLS code generated from hls4ml is passed to HLS backends, and merge with the RTL flow to generate the FPGA bitstream with synthesis and place and route backends.}
\end{figure*}

\subsection{FPGA resource consumption estimation}
\label{sec:mul}
On-chip resource consumption of neural networks is often dominated by matrix-vector multiplication operations.
A common metric for estimating resource usage of such operations on FPGAs is Bit Operations (BOPs)~\cite{uniq}.
However, we find that BOPs often significantly overestimates the resource consumption for unrolled logic.
We propose a new metric, Effective Bit Operations (EBOPs), to accurately estimate the resource consumption of  CMVM operations in models on FPGAs.
The total EBOPs of a model is given by
\begin{equation}
    \mathrm{EBOPs} = \sum_{{i,j}\in\mathcal{M}} b_\mathrm{i}\cdot b_\mathrm{j} + \sum_{{k,l}\in\mathcal{A}}\max(b_k,b_l),
\end{equation}
where $\mathcal{M} = \left\{\{i,j\}_n\right\}$ is the set of all multiplication operations between operands with bit-widths $b_i$ and $b_j$ in that model, and $\mathcal{A} = \left\{\{k,l\}_m\right\}$ is the set of all explicit addition/subtraction/muxing operations (e.g., bias, average or maximum pooling) between operands with bit-widths $b_k$ and $b_l$.
In practice, EBOPs is typically dominated by the first term, which is constructed based on a simplified model of the resource consumption of a constant multiplier on FPGAs: We assume that the multiplication of a constant (e.g., a weight) and a variable (e.g., an activation) is implemented as a series of shift-and-add operations.
Let the bit-width of the variable and the constant be $b_v$ and $b_c$ (excluding sign), respectively.
The multiplication can be implemented with $b_c - 1$ shift-and-add operations in the worst case, such that resource consumption is linear in $b_v\cdot b_c$ when the result is further accumulated with other multiplications in a CMVM operation.
This relation can still be used if canonical signed digits~\cite{csd} representation is adopted for the constants when using the \texttt{da4ml} backend, where EBOPs is proportional to the number of signed digits weighted by the bit-width of the variable operand.
In this case, EBOPs can be regarded as a surrogate for the initial complexity of the CMVM operations before common subexpression elimination is applied, and we find that it still maintains a strong correlation with resource consumption after place-and-route.

Experimental findings confirm that EBOPs is a reliable estimator for on-chip resource consumption when using naive unrolled implementations in \texttt{hls4ml}, which closely mirrors a linear combination of usage of look-up tables (LUTs) and usage of digital signal processors (DSPs).
In \texttt{da4ml}, the direct linear relation breaks down due to further optimizations performed, but EBOPs maintains a strong positive correlation, so it is still a useful differentiable surrogate.

EBOPs is incorporated in the loss function as a regularization term with a coefficient $\beta\in\mathbb{R}_+$ to obtain a trade-off between performance and on-chip resource usage.
Since some network parameters are not involved in multiplication operations, e.g., the last-layer's outputs or inputs, we add another $L_1$ regularization with a coefficient $\gamma\in\mathbb{R}_+$ to the bit-widths to keep them from growing indefinitely.
Hence, the final loss function is given by
\begin{equation}
    \mathcal{L} = \mathcal{L}_\mathrm{base} + \beta\cdot\mathrm{EBOPs} + \gamma\cdot \sum\text{bit-widths},
\end{equation}

As all additional gradients introduced in this section only apply to bit-widths, the loss landscape of the network weights remains unperturbed compared to that of networks with static quantization parameters.

\section{Implementation}

The HGQ method is implemented as an easy-to-use, standalone Python package, \texttt{HGQ}, on top of Keras v3~\cite{keras}.
The package is open-source under the LGPL-3.0 license.

While the PyTorch style, dynamic graph-based deep learning frameworks are popular in the research community, we choose Keras as the base framework since dynamic graph features cannot be easily supported when exporting the models to RTL/HLS for FPGA deployment.
In contrast, Keras uses a static graph-based framework\footnote{The underlying graph execution engines support dynamic graphs, while the Keras front-end itself makes a static graph when using \texttt{keras.Model} by default.}, which provides a  straightforward path for training and exporting the models to synthesizable RTL/HLS designs.
Furthermore, since Keras v3 supports multiple backends, \texttt{HGQ} can be used with all Keras-supported backends, including JAX~\cite{jax}, TensorFlow~\cite{tensorflow}, and PyTorch~\cite{torch}.
When using the PyTorch backend, one could still directly use Keras layers as torch modules, so Keras is a suitable choice for implementing \texttt{HGQ}.
Depending on the specific training condition, HGQ's training speed is typically $50\%$ to $90\%$ of that of the corresponding native Keras models when using the JAX or TensorFlow backends, and $30\%$ to $80\%$ when using the PyTorch backend. In practice, this means that any model that can be trained with the underlying framework can also be trained with HGQ at comparable cost, so there is no intrinsic limitation on model size from the training side.

Specifically, XLA compilation provided by JAX and TensorFlow is supported and can provide significant speedup during training, especially for small models aiming for FPGA deployment.
In addition to standard GPU training, \texttt{HGQ} also supports training on Google TPU~\cite{tpuv4} when using the JAX or TensorFlow backend.

Although our evaluation focuses on dense, convolutional, and graph neural networks, \texttt{HGQ} supports more complex operations, including multihead attention layer~\cite{transformer} and Linformer attention layer~\cite{linformer}, which can be used to construct and train Transformer or Linformer models with heterogeneous quantization.

We add support for both \texttt{da4ml} and \texttt{hls4ml} backends to import models trained with \texttt{HGQ}, and convert them into synthesizable RTL/HLS bit-exact designs.
Partial support is also added to the \texttt{QONNX} project~\cite{qonnx} to import \texttt{HGQ} models.

The overall workflow of using HGQ is shown in Fig.~\ref{fig:workflow}.
A user first constructs a model with HGQ-provided quantized layers in the same way as the original Keras model, and then trains the model like any other Keras models in their preferred manner.
After training, the user may convert the model into a synthesizable RTL/HLS design with either the \texttt{da4ml} or \texttt{hls4ml} backend, and follow the respective workflow to obtain an RTL design to integrate into their FPGA implementation.
While heterogeneous activation quantization is not supported, users may choose to export the models to QONNX format for other backends, such as \texttt{FINN}.

\section{Evaluation and Analysis}
\label{sec:results}

\subsection{Experimental Setup}
\subsubsection{Task description and model architectures}
To evaluate the performance of HGQ, we train and evaluate models on a variety of tasks, including the jet substructure classification (JSC) task at the LHC~\cite{jet_dataset}, the SVHN image classification task~\cite{svhn}, and the muon tracking task~\cite{tgc}.
For the JSC task, two variants of inputs for the same task are available: one with 16 high-level features (HLF), and the other with up to 150 particle-level features (PLF).
For the HLF inputs, two versions of the dataset are available: one hosted on OpenML~\cite{openml-jet}, and the other hosted on CERNBox~\cite{cernbox-jet}.

All JSC tasks are classification tasks with 5 classes, and the performance is evaluated using the test accuracy.
For both HLF JSC tasks, we use a fully-connected network with 3 hidden layers taken from~\cite{qkeras}, and the size for the datalanes is \texttt{16-64-32-32-5}.
For the PLF JSC task, the input is an $N\times n$ array, where $N$ is the maximum number of particles, and $n$ is the number of features per particle.
The array is padded with zeros when the number of particles is less than $N$.
When $n=3$, we apply an additional selection of $p_\mathrm{T}>2$\footnote{The $p_\mathrm{T}$ is the transverse momentum of the particle with respect to the beam axis.} GeV in align with~\cite{ds-fpga}.
We use a linearized graph neural network, JEDI-linear~\cite{jedi-linear}, for all PLF JSC tasks.
The SVHN task is a 10-class classification task with an input image of size $32\times32\times3$.
We use a LeNet-like~\cite{lenet}, small CNN with 3 convolutional layers and 2 fully-connected layers taken from~\cite{fast_cnn}.
The muon tracking task is a regression task to predict the angle of incidence of a muon passing through a simplified model of the ATLAS Thin Gap Chambers (TGC) detector~\cite{tgc}~\cite{atlas-tdr-029}.
The input is a set of three 2-d arrays of size $50\times3$, $50\times2$, and $50\times2$, representing the three stations of the TGC detector, respectively.
All arrays are binary-valued, with 1 representing a hit in that pixel.
The model used is taken from~\cite{tgc}, which is a multi-stage model designed to align with the dynamics of the detector. The exact architectures for the models used for these tasks can be found in their respective references.
The performance is evaluated with the test resolution, defined as the root-mean-square of the residual between the predicted and true angle of incidence, excluding up to 0.1\% of outliers greater than 30 mrad away from the true value.

All resource/latency values in this section from the HGQ-trained models are obtained after place-and-route out-of-context with Vivado 2025.1 targeting Xilinx Virtex UltraScale+ VU13P FPGA\footnote{One of the two (the other one is VU9P) proposed FPGA models for the majority of both ATLAS and CMS Experiments' future trigger system, which is also a common benchmark target used in prior works.} (part: \texttt{xcvu13p-flga2577-2-e}) with global retiming enabled.
The model accuracy is measured on the generated HLS or RTL models over test datasets.
For all RTL models generated by \texttt{da4ml}, we verify the functional correctness with Verilator~\cite{verilator} against the Keras models.
For all models obtained, we did not notice any bit-mismatches between the Verilated models and the Keras models that cannot be explained by floating-point rounding errors, and there is no noticeable\footnote{$>99.5$\% of the models we tested had 0 bit mismatch on their corresponding test datasets.
    For those models with at least one mismatch, no larger than $10^{-3}$\% accuracy degradation is observed.
    Only models requiring at least 23 bits during accumulation may have mismatches when using the JAX backend running on CPU.} difference in the accuracy/resolution between the Keras models and the Verilated models.

\begin{table}
    \centering
    \begin{adjustbox}{width=0.45\textwidth}
        \begin{tabular}{llccc}
            \toprule
            Task                                & Batch size & HGQ   & Keras & NLA  \\
            \midrule
            HLF JSC                             & 16600      & 0.645 & 0.414 & 164. \\
            JSC PLF (32 particles, 16 features) & 2790       & 1.85  & 1.25  & -    \\
            JSC PLF (64 particles, 16 features) & 2790       & 4.10  & 2.76  & -    \\
            TGC Muon Tracking                   & 51200      & 2.30  & 1.68  & -    \\
            SVHN Classifier                     & 2048       & 5.03  & 3.49  & -    \\
            \bottomrule
        \end{tabular}
    \end{adjustbox}
    \caption{Training time of HGQ, native Keras, and NeuraLUT-Assemble (NLA) on various tasks. Results are obtained on a single NVIDIA RTX 4090 GPU with Core i7 13700 CPU. The unit is in milliseconds (ms) per step at specified batch size.}
    \label{tab:training_speed}
    \vspace{-0.8cm}
\end{table}

\begin{figure}
    \centering
    \includegraphics[width=0.45\textwidth]{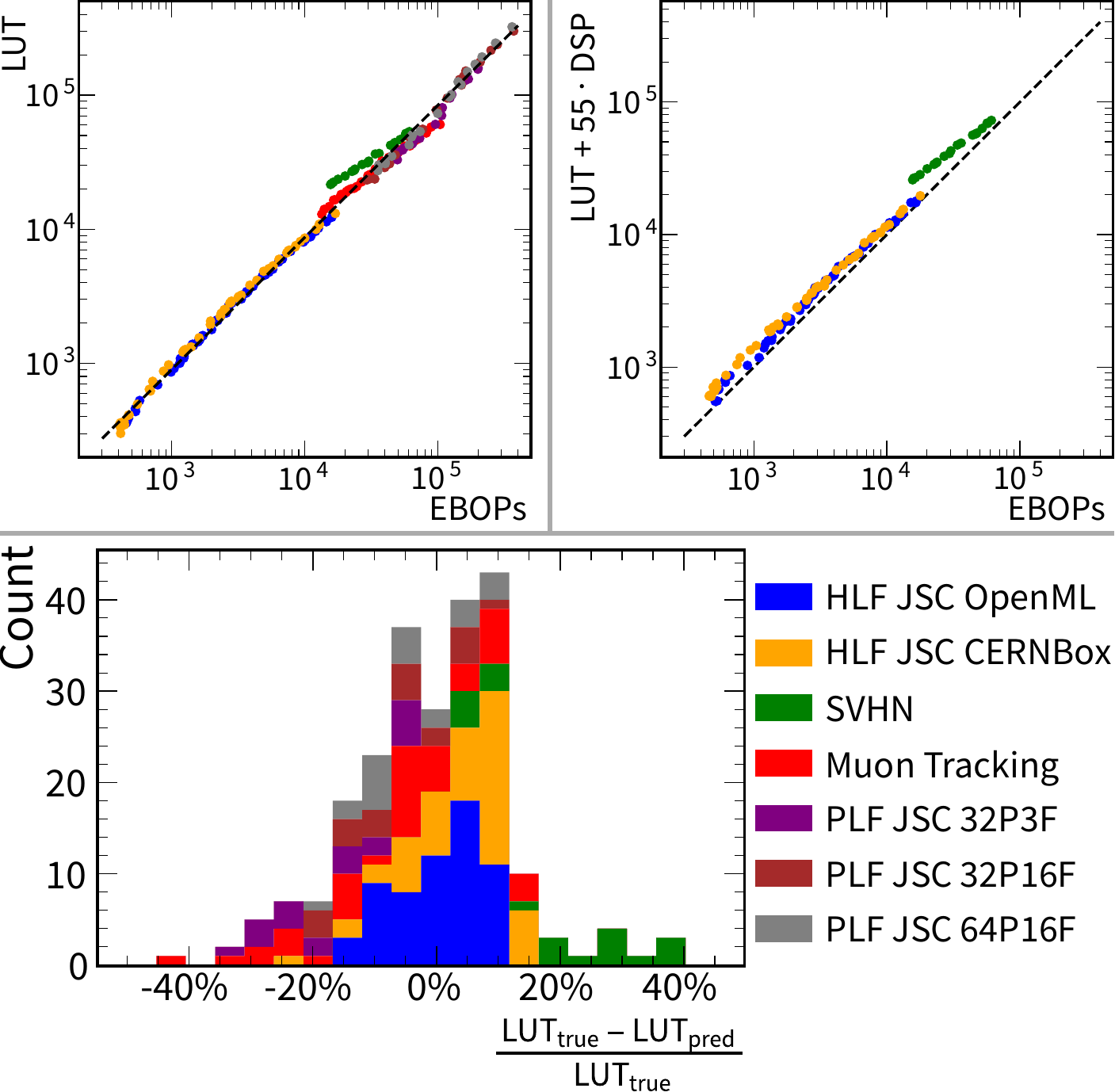}
    \caption{The relationship between EBOPs and the post place-and-route resource consumption with (top left) and without (top right) DA optimization, and the distribution of error between the estimated and actual LUT consumption after place-and-route with DA (lower). For the error distribution, $\mathrm{LUT}_\mathrm{pred}$ is given by $\exp(0.985\cdot\log(\mathrm{EBOPs}))$.
    }
    \label{fig:ebops}
    \Description{
        Three plots validating Estimated Binary Operations against resource usage for various machine learning and tracking datasets.

        The figure consists of two scatter plots at the top and a stacked histogram at the bottom. The top-left log-log scatter plot shows Look Up Tables versus Estimated Binary Operations, ranging from 10 to the power of 3 to 5 times 10 to the power of 5. Data points for all datasets closely follow a diagonal dashed line, indicating a high correlation between the predicted and true values.

        The top-right log-log scatter plot shows the sum of Look Up Tables and 55 times Digital Signal Processors versus Estimated Binary Operations. This plot maintains the linear correlation, though the Street View House Numbers dataset points are clustered slightly above the diagonal line compared to other datasets.

        The bottom chart is a stacked histogram showing the count of relative error, defined as the difference between true and predicted Look Up Tables divided by true Look Up Tables. The horizontal axis represents percentage error from negative 40 percent to positive 40 percent. The total distribution peaks between 0 and 15 percent error. High Level Feature OpenML and CERNBox datasets, along with Muon Tracking, are clustered primarily between negative 10 and positive 15 percent. The Street View House Numbers dataset shows a distinct spread toward positive error, ranging from 10 to 40 percent. Parallel Linear Filter datasets at various precisions, including 32-bit and 64-bit configurations, are distributed mostly between negative 30 percent and 10 percent error.
    }
\end{figure}

\subsubsection{Training configuration}

To explore the trade-off between accuracy (or resolution) and resource usage with different quantization levels, we adjust the value of the $\beta$ factor for each task during one training session to map out the Pareto fronts. For each training session, we initialize the model with a small $\beta$ value and then gradually increase it throughout the training with an exponential schedule. The initial and final $\beta$ values vary depending on the task: \texttt{5e-7} to \texttt{1e-3} for HLF JSC, \texttt{2e-8} to \texttt{3e-6} for both PLF JSC and TGC Muon Tracking, and \texttt{1e-7} to \texttt{1e-5} for SVHN Classification. Meanwhile, we maintain the $\gamma$ value fixed at an arbitrarily small value of \texttt{2e-8} for all experiments to avert the risk of diverging bit-widths for some parameters not involved in multiplication operations.

We use the Adam optimizer~\cite{adam} with a cosine annealing with restart learning rate schedule for all experiments.
After each epoch, we record the validation accuracy and EBOPs, and save the checkpoints for models that are on the Pareto front defined by these two metrics for further evaluation.
Post-training, we use the entire training and validation datasets for calibration to determine the required bit-widths and evaluate the exact EBOPs for all the checkpointed models.

The training time per step for HGQ, native Keras, and NeuraLUT-Assemble~\cite{neuralut-assemble} are compared in Tab.~\ref{tab:training_speed}. For the models benchmarked, HGQ achieves a training speed that is $\sim70\%$ of the native Keras models, whereas NeuraLUT-Assemble is significantly slower ($\sim254$ times slower for the HLF JSC task).

\begin{figure}
    \centering
    \includegraphics[width=0.45\textwidth]{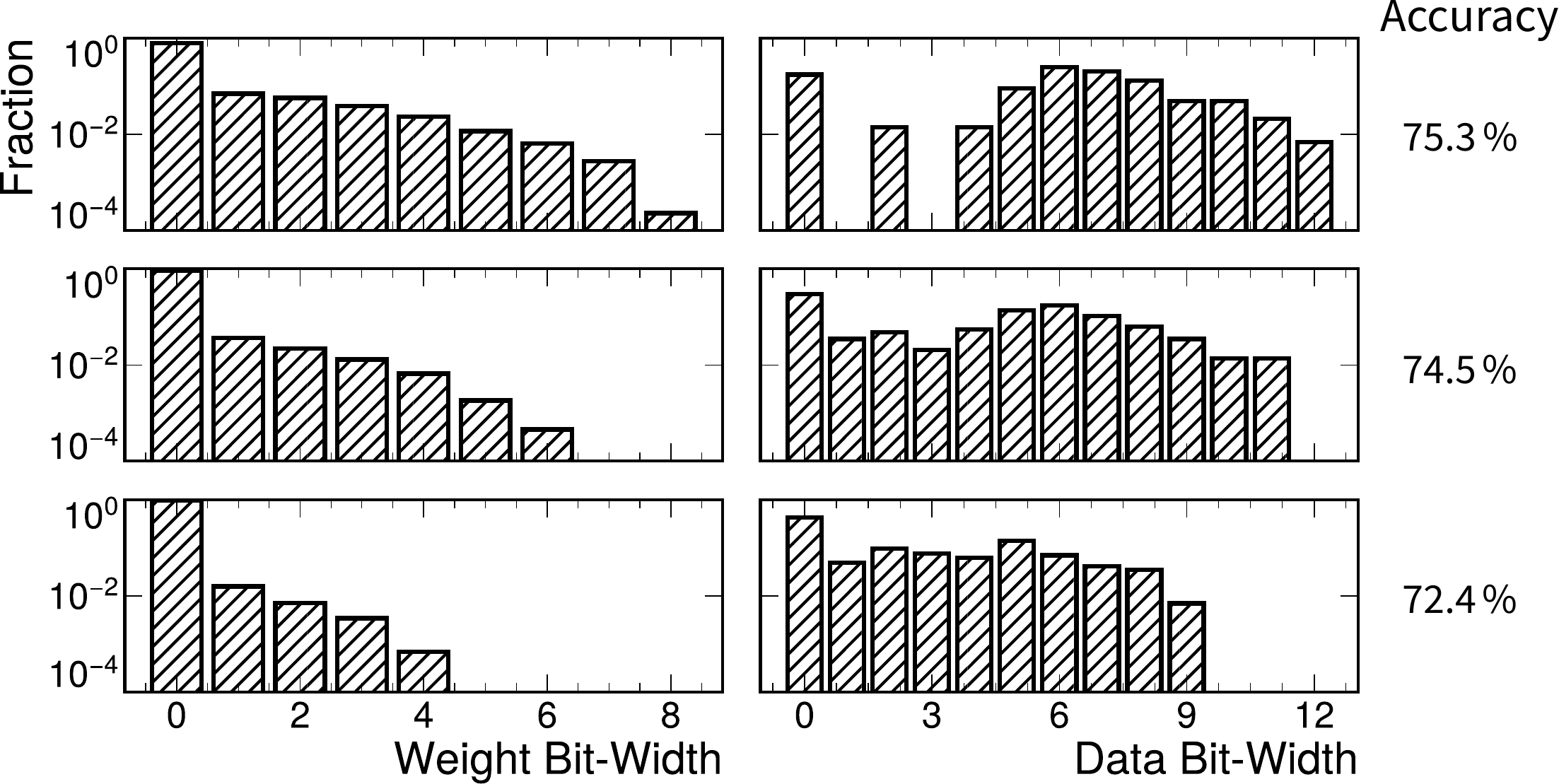}
    \caption{Distributions of the weights and data activations bit-widths of the HGQ models for HLF JSC task, CERNBox version.}
    \label{fig:bw_dist_cernbox}
    \vspace{-0.4cm}
    \Description{
        The figure consists of a three-by-two grid of bar charts. Each row represents a model configuration with a specific accuracy labeled on the far right: 75.3 percent for the top row, 74.5 percent for the middle row, and 72.4 percent for the bottom row. The vertical axis for all charts represents the fraction of elements on a logarithmic scale ranging from 10 to the power of negative 4 to 1.

        The left column displays Weight Bit-Width distributions. In all three cases, the largest fraction is at 0 bits. As accuracy decreases from top to bottom, the maximum weight bit-width used decreases from 8 bits to 6 bits and finally to 4 bits.

        The right column displays Data Bit-Width distributions. For the 75.3 percent and 74.5 percent accuracy models, the data bit-widths range from 0 to 12. For the 72.4 percent accuracy model, the distribution is more restricted, ranging from 0 to 9 bits.

        For both weights and data bit-width distributions, the overall distribution shifts towards zero as accuracy decreases.
    }
\end{figure}

\begin{figure}
    \centering
    \includegraphics[width=0.45\textwidth]{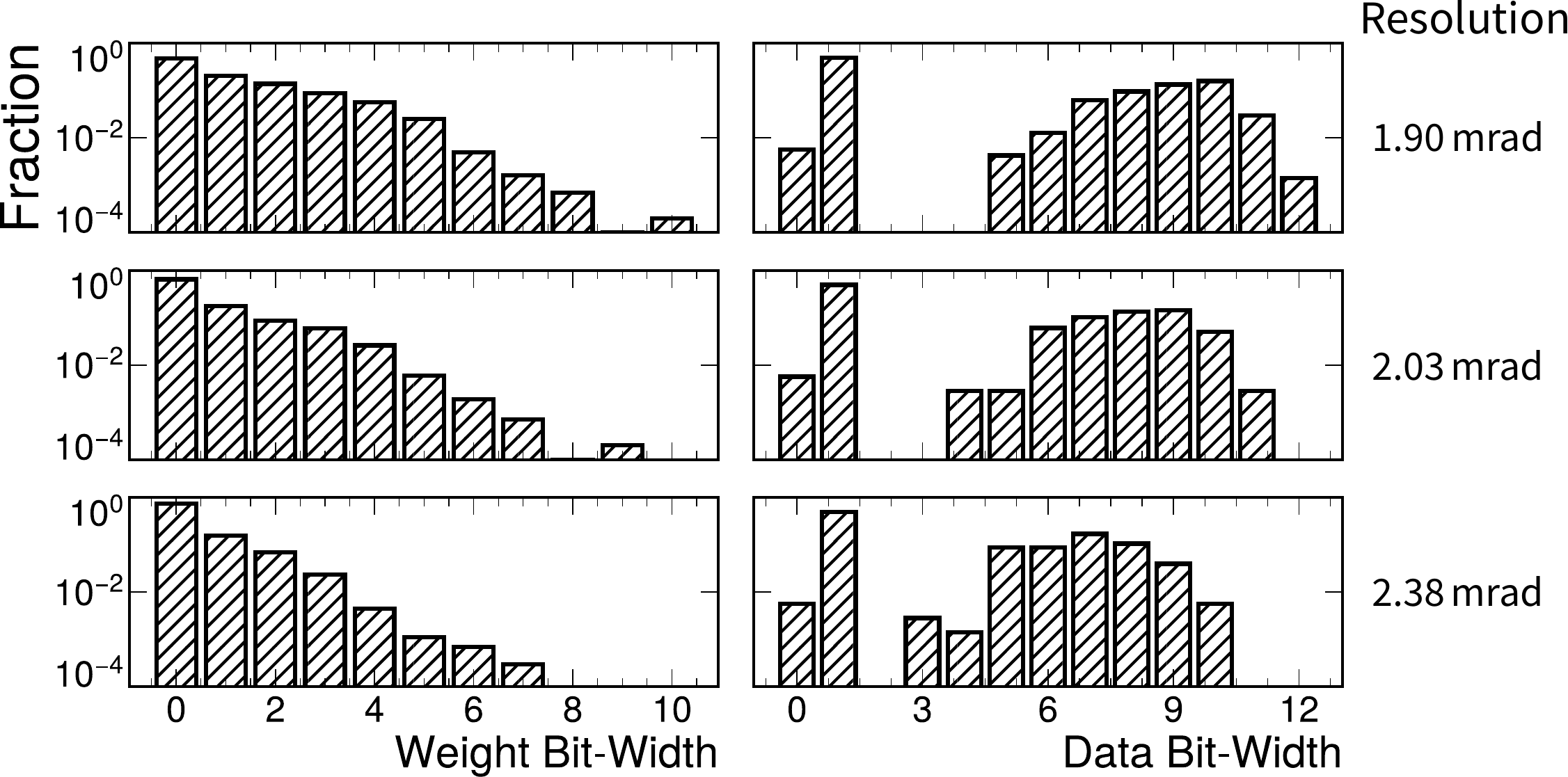}
    \caption{Distributions of the weights and data activations bit-widths of the HGQ models for the muon tracking task.}
    \label{fig:bw_dist_tgc}
    \vspace{-0.3cm}
    \Description{
        Similar to the other bit-width figure, this figure shows a three-by-two bar chart, showing decreasing bit-widths of the weights and data as the performance of the muon tracking network decreasing.
    }
\end{figure}

\import{.}{/tables/hlf_jsc.tex}
\import{.}{/tables/plf_jsc.tex}
\import{.}{/tables/tgc_svhn.tex}

\subsection{Resource Estimation via EBOPs}

We first demonstrate that EBOPs is a reasonable differentiable surrogate for on-chip resource consumption with both \texttt{da4ml} and \texttt{hls4ml}. Empirically, we find that EBOPs maintains a strong correlation with the actual on-chip resource consumption after place-and-route when using \texttt{da4ml} as the backend for each individual architecture, as shown on the top left of Fig.~\ref{fig:ebops}.
Within each architecture, as EBOPs increases, the LUT consumption also increases monotonically with a diminishing rate.
This diminishing rate is expected as common subexpression elimination used in \texttt{da4ml} is effective when the CMVM problems are more complex and with higher bit-widths.
While the consumption for each individual architecture is not linear to EBOPs, the overall trend combining different architectures is approximately linear in logarithmic space with a very small intercept, and an approximate surrogate relation of $\mathrm{LUT} \approx \exp(0.985\cdot\log(\mathrm{EBOPs}))$ can be used to estimate LUT consumption from EBOPs.
The error distribution between estimated and actual LUT consumption is shown on the lower plot of Fig.~\ref{fig:ebops}, where the relative error is within 20\% for the majority of the models, and the actual LUT consumption never exceeds more than 20\% of the estimated value among all models we tested, excluding the SVHN Classifier model which requires significant additional switching and buffering logic not modeled by EBOPs.
Hence, EBOPs could be a useful rough estimator for model selection, for instance, during model architecture optimizations.
While EBOPs does not provide the most accurate resource estimation by itself, as shown on the top left of Fig.~\ref{fig:ebops}, the relative ordering of resource consumption is mostly preserved -- for each individual architecture, a model with higher EBOPs always consumes more LUTs than a model with lower EBOPs when implemented in the same way.
This suggests that EBOPs is a reasonable differentiable surrogate for LUT consumption when using \texttt{da4ml} as the backend.

When using \texttt{hls4ml} as the backend, we find EBOPs to be approximately a linear combination of LUT and DSP consumption: $\mathrm{EBOPs}\approx \mathrm{LUT} + 55\cdot\mathrm{DSP}$, as shown on the top right of Fig.~\ref{fig:ebops}.
This relation is also empirically observed for the Xilinx UltraScale+ chips, and it may be different for other series of FPGAs or other vendors.

However, it is important to note that EBOPs only estimates resource consumption from the CMVM operations and other arithmetic operations in the network.
Other non-arithmetic operations that are not directly associated with the network architecture, such as those arising from the control logic for processing element scheduling, or FIFOs for buffering added at implementation time when needed, are not currently accounted for in EBOPs.
In these cases, the actual resource consumption will be higher than what EBOPs predicts. For example the SVHN classifier, which requires additional FIFO buffers and intensive muxing logic that are not modeled in EBOPs, is shown on the lower plot of Fig.~\ref{fig:ebops}.
We validate that the resulting models have heterogeneous bit-widths as expected, and we show respectively the distribution of the bit-widths for the weights and activations of a few representative models in Fig.~\ref{fig:bw_dist_cernbox} and Fig.~\ref{fig:bw_dist_tgc} for the HLF JSC CERNBox and the muon tracking tasks.
The bit-widths are shown to distribute across a wide range, with the majority of the weights being quantized to zero bits (pruned), and the bit-widths becoming smaller in general as the model performance is traded off for lower resource consumption.

\subsection{Comparison with Prior Work}

\subsubsection{HLF JSC tasks}
\begin{figure}
    \centering
    \includegraphics[width=0.45\textwidth]{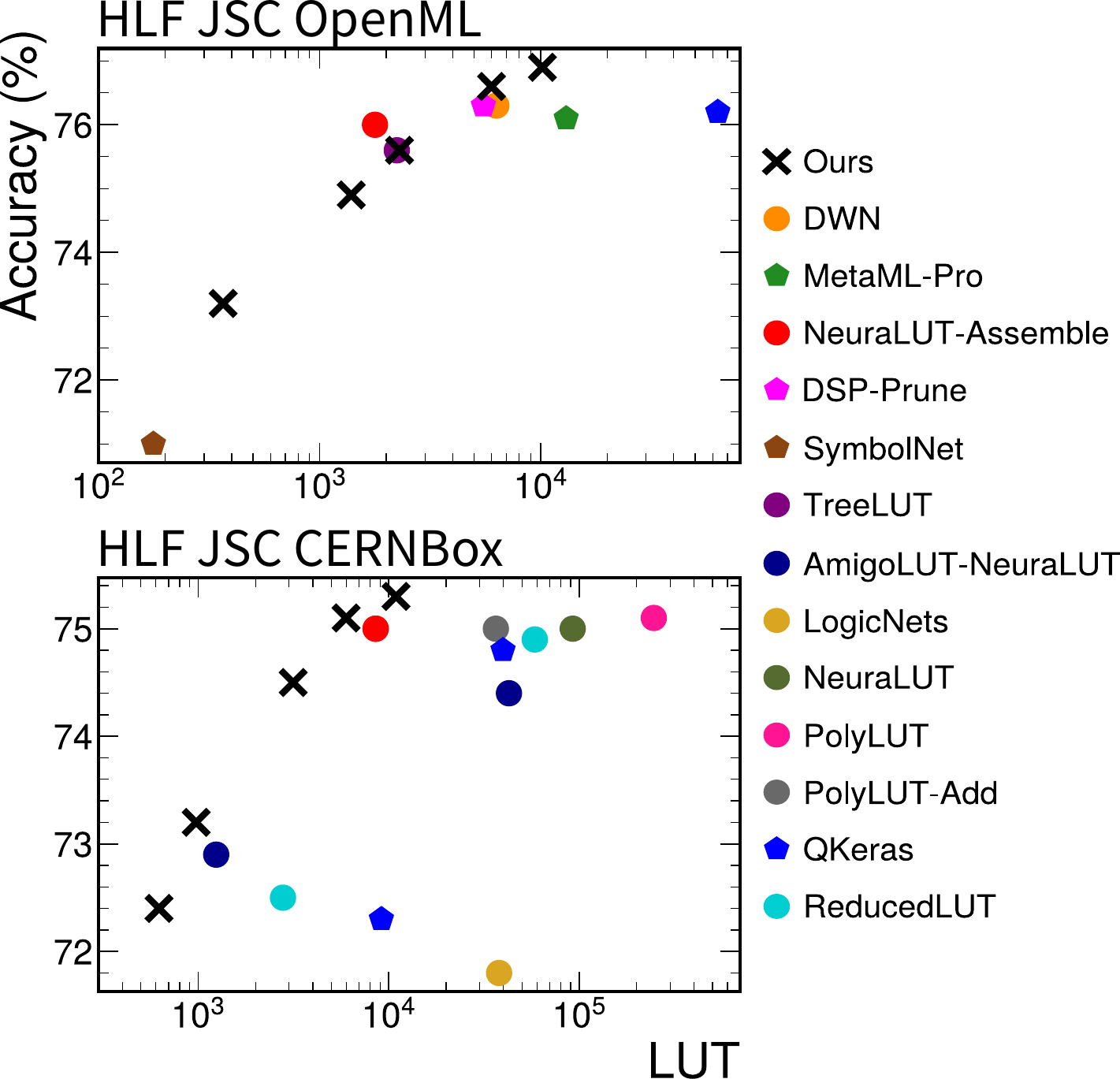}
    \caption{
        Accuracy versus LUT consumption of the HLF JSC models on the OpenML and CERNBox  datasets.
        Designs with circle markers use only LUTs for logical operations, while designs with pentagon markers use both LUTs and DSPs.
        Our designs are marked as "X", and uses only LUTs.}
    \label{fig:hlf_jsc_plt}
    \vspace{-0.6cm}
    \Description{
        This figure is a visualization of the LUT consumption and accuracy obtained for each method shown in table 3.
    }
\end{figure}

We compare the performance of the models trained with HGQ against various prior efforts, including quantized neural networks, LUT-based models, decision forests, and symbolic models on the HLF JSC task.
The accuracy, latency, \fmax, and on-chip resource utilization of various models are summarized in Tab.~\ref{tab:hlf_jsc}.
To visualize the trade-off between accuracy and resource consumption, we also study accuracy versus LUT consumption in Fig.~\ref{fig:hlf_jsc_plt} for the two HLF JSC tasks.
In both datasets, the HGQ-trained models outperform all previous methods based on quantized neural networks by a large margin, both in terms of accuracy and resource usage, and achieve similar accuracy-resource trade-offs as the state-of-the-art decision forest and symbolic models.

Compared to NeuraLUT-Assemble~\cite{neuralut-assemble} on the OpenML dataset, HGQ can achieve higher accuracy, though achieving the same accuracy typically consumes more resources. On CERNBox, HGQ outperforms NeuraLUT-Assemble in both accuracy and resource use. Latency-wise, HGQ is substantially faster than prior quantized neural networks but remains slower than NeuraLUT-Assemble. For simple classification tasks that fit within a few thousand LUTs, both approaches are competitive. But, LUT-based methods have not yet been shown to scale to models with tens to hundreds of thousands of LUTs, which are often required for more complex problems. HGQ preserves standard neural-network semantics and scales to deeper/wider models and to a broader set of architectures (beyond MLPs), as shown in Tab.~\ref{tab:plf_jsc} and Tab.~\ref{tab:tgc_svhn}.
HGQ can also utilize both LUT and DSP resources (Tab.~\ref{tab:tgc_svhn}), and it has a significantly shorter training time than competing approaches (Tab.~\ref{tab:training_speed}).

HGQ has been smoothly integrated into \texttt{da4ml} and \texttt{hls4ml} which makes it vendor-portable, whereas current LUT-based approaches are largely tied to AMD devices.
Although LUT-based pipelines can deliver the very lowest latencies at high accuracy, level-1 trigger applications do not necessarily require such extreme margins since the LHC collision occurs every 25 ns, so HGQ can produce competitive end-to-end designs.
These properties give HGQ better scalability, architectural generality, heterogeneous-resource efficiency, and cross-vendor portability, making it a more versatile choice for larger networks and broader deployments.
As noted in related work, HGQ can also be combined with NeuraLUT-Assemble to further improve hardware efficiency; we leave this integration to future work.

\subsubsection{PLF JSC tasks}
We show the performance of  HGQ-trained models for the PLF JSC tasks in Tab.~\ref{tab:plf_jsc}, where the input contains up to 32 or 64 particles with 3 or 16 features per particle, respectively.
On this task, HGQ outperforms the prior quantized neural network based models on both accuracy and resource consumption by a substantial margin. We also show an MLP-Mixer (MLPM) based model~\cite{mlpm} design running on an Altera FPGA, showing that HGQ can be applied to FPGAs from other vendors other than AMD.

\subsubsection{Other tasks}
As shown in Tab.~\ref{tab:tgc_svhn},  the HGQ-trained models outperform the prior art on the muon tracking task with significant resource reduction and speedup while achieving higher accuracy.
On the SVHN classification task, since intense resource reuse is required to fit the model into the FPGA, the latency is significantly higher than other models so we use the \texttt{hls4ml} backend for this task.
The resulting models achieve similar accuracy while using less resource, compared to prior quantized neural network based models on this task. It also shows that our approach can use both LUT and DSPs efficiently.

\subsection{Ablation}

To isolate the contribution of HGQ from backend optimizations in~\texttt{da4ml} (e.g., common subexpression elimination), we conduct an ablation study to show the effectiveness of HGQ when using \texttt{hls4ml} as the backend without DA adoption.
We show the same models for the HLF JSC tasks converted by \texttt{hls4ml} and synthesized with Vivado 2025.1 in Tab.~\ref{tab:da_v_hls4ml} with prior works using \texttt{hls4ml} as the backend for comparison.
Compared with the prior quantized neural network based models, HGQ consistently delivers superior accuracy-efficiency trade-offs also on the \texttt{hls4ml} backend, achieving higher accuracy with lower DSP, LUT consumption and achieves lower latency at a much higher \fmax.

\section{Conclusion and Outlook}

This paper presents HGQ, a novel method for training neural networks with heterogeneous quantization for efficient FPGA deployment.
The method is validated on a variety of tasks and shows significant improvement over prior efforts on quantized neural networks,  achieving competitive performance with state-of-the-art alternative methods on classification tasks.
The method is implemented in \texttt{HGQ}, an easy-to-use, open-source Python package which is compatible with multiple backends and supports exporting to synthesizable bit-exact RTL/HLS designs.
Future work includes extending the optimization methodology used in HGQ to LUT-based methods, such as NeuraLUT or NeuraLUT-Assemble. Since these methods require quantization on the sub-network level, HGQ could be used potentially to further enhance the hardware efficiency. In addition, HGQ could also be used to produce different neural architectures for them.
Moreover, while more complex models such as transformers are supported in \texttt{HGQ}, more work is required for efficient FPGA deployment of these models, and we are working with the \texttt{hls4ml} team on optimizing them for FPGA implementations.

\import{.}{/tables/da_v_hls4ml.tex}

\nocite{gnu_parallel}

\section*{Acknowledgement}
Partial support from the United States DoE (grant numbers DE-SC0011925, DE-FOA-0002705), NSF (grant numbers PHY240298, PHY2117997), United Kingdom EPSRC (grant numbers UKRI256, EP/V028251/1, EP/N031768/1, EP/S030069/1, and EP/X036006/1), Swiss NSF Grant No.~PZ00P2\_201594, Schmidt Futures (Grant G-23-64934), KIAT, Intel, and AMD is acknowledged. We acknowledge the Caltech Danny Koh graduate student scholarship and the ETH/Guenther Dissertori partial support for this project.
\bibliographystyle{ACM-Reference-Format}
\balance
\bibliography{bibliography}

\end{document}

%% file: tables/hlf_jsc.tex
\begin{table*}[htbp]
    \begin{adjustbox}{width=0.9\textwidth,center=\textwidth}
        \setlength\extrarowheight{-0.8pt}
        \begin{tabular}{l|ccccccc}
            \multicolumn{8}{c}{\textbf{HLF JSC (OpenML)}}                                                                                                               \\
            \midrule
            Implementation                                                & Accuracy $\uparrow$ & Latency [cycles] & LUT     & DSP & FF     & \fmax [MHz] & II [cycles] \\
            \midrule
            \textbf{HGQ}                                                  & 76.9\%              & 20 (36.3 ns)     & 10,182  & 0   & 10,480 & 551.6       & 1           \\
            \textbf{HGQ}                                                  & 76.6\%              & 16 (26.2 ns)     & 5,991   & 0   & 5,890  & 611.2       & 1           \\
            \textbf{HGQ}                                                  & 75.6\%              & 12 (18.6 ns)     & 2,298   & 0   & 2,217  & 645.2       & 1           \\
            \textbf{HGQ}                                                  & 74.9\%              & 10 (14.5 ns)     & 1,390   & 0   & 1,316  & 691.6       & 1           \\
            \textbf{HGQ}                                                  & 73.2\%              & 6 (9.1 ns)       & 366     & 0   & 363    & 662.7       & 1           \\
            \texttt{QKeras} [ICFPT'23]~\cite{dsp-prune}$^{**}$            & 76.3\%              & 15 (105.0 ns)    & 5,504   & 175 & 3,036  & $>142.9$    & 2           \\
            \texttt{DWN} [ICLR'24]~\cite{dwn}$^{***}$                     & 76.3\%              & 10 (14.4 ns)     & 6,302   & 0   & 4,128  & 695.        & 1           \\
            \texttt{QKeras} [CoRR'21]~\cite{hls4ml}$^{*}$                 & 76.2\%              & 9 (45)           & 63,251  & 38, & 4,394  & $\sim 200$  & 1           \\
            \texttt{MetaML-Pro} [TRETS'26]~\cite{metamlpro}$^{*}$         & 76.1\%              & 10 (50 ns)       & 13,042  & 70  & N/A    & $\sim200$   & 1           \\
            \texttt{NeuraLUT-Assemble} [FCCM'25]~\cite{neuralut-assemble} & 76.0\%              & 2 (2.1 ns)       & 1,780   & 0   & 540    & 940.        & 1           \\
            \texttt{TreeLUT} [FPGA'25]~\cite{treelut}                     & 75.6\%              & 2 (2.7 ns)       & 2,234   & 0   & 347    & 735.        & 1           \\
            \texttt{SymbolNet} [MLST'25]~\cite{symbolnet}                 & 71.\%               & 2 (10 ns)        & 177     & 3   & 109    & $\sim200$   & 1           \\
            \midrule

            \multicolumn{8}{c}{\textbf{HLF JSC (CERNBox)}}                                                                                                              \\
            \midrule
            Implementation                                                & Accuracy $\uparrow$ & Latency [cycles] & LUT     & DSP & FF     & \fmax [MHz] & II [cycles] \\
            \midrule
            \textbf{HGQ}                                                  & 75.3\%              & 18 (31.1 ns)     & 10,921  & 0   & 11,183 & 578.4       & 1           \\
            \textbf{HGQ}                                                  & 75.1\%              & 15 (24.6 ns)     & 5,974   & 0   & 5,775  & 609.8       & 1           \\
            \textbf{HGQ}                                                  & 74.5\%              & 13 (20.4 ns)     & 3,152   & 0   & 2,941  & 636.9       & 1           \\
            \textbf{HGQ}                                                  & 73.2\%              & 11 (14.6 ns)     & 976     & 0   & 904    & 754.1       & 1           \\
            \textbf{HGQ}                                                  & 72.4\%              & 9 (9.9 ns)       & 623     & 0   & 642    & 905.8       & 1           \\
            \textbf{HGQ}$^\dagger$                                        & 72.4\%              & 2 (5.4 ns)       & 641     & 0   & 227    & 367.4       & 1           \\
            \texttt{PolyLUT} [TC'25]~\cite{polylut}                       & 75.1\%              & 5 (24.6 ns)      & 246,071 & 0   & 12,384 & 203.        & 1           \\
            \texttt{NeuraLUT-Assemble} [FCCM'25]~\cite{neuralut-assemble} & 75.0\%              & 2 (5.7 ns)       & 8,539   & 0   & 1,332  & 352.        & 1           \\
            \texttt{NeuraLUT-Assemble} [FCCM'25]~\cite{neuralut-assemble} & 75.0\%              & 7 (7.0 ns)       & 8,535   & 0   & 2,717  & 994.        & 1           \\
            \texttt{PolyLUT-Add} [FPL'24]~\cite{polylut-add}              & 75.\%               & 5 (15.9 ns)      & 36,484  & 0   & 1,209  & 315.        & 1           \\
            \texttt{NeuraLUT} [FPL'24]~\cite{neuralut}                    & 75.\%               & 5 (13.6 ns)      & 92,357  & 0   & 4,885  & 368.        & 1           \\
            \texttt{ReducedLUT} [FPGA'25]~\cite{reducedlut}               & 74.9\%              & N/A              & 58,409  & 0   & N/A    & 302.8       & N/A         \\
            \texttt{ReducedLUT} [FPGA'25]~\cite{reducedlut}               & 72.5\%              & N/A              & 2,786   & 0   & N/A    & 408.5       & N/A         \\
            \texttt{QKeras} [NMI'21]~\cite{qkeras}$^{*}$                  & 74.8\%              & 11 (55 ns)       & 39,782  & 124 & 8,128  & $\sim 200$  & 1           \\
            \texttt{QKeras} [NMI'21]~\cite{qkeras}$^{*}$                  & 72.3\%              & 11 (55 ns)       & 9,149   & 66  & 1,781  & $\sim 200$  & 1           \\
            \texttt{AmigoLUT-NeuraLUT} [FPGA'25]~\cite{amigolut}          & 74.4\%              & 5 (9.6 ns)       & 42,742  & 0   & 4,717  & 520.        & 1           \\
            \texttt{AmigoLUT-NeuraLUT} [FPGA'25]~\cite{amigolut}          & 72.9\%              & 5 (5.0 ns)       & 1,243   & 0   & 1,240  & 1,008.      & 1           \\
            \texttt{LogicNets} [FPL'20]~\cite{logicnets}                  & 71.8\%              & 5 (11.7 ns)      & 37,931  & 0   & 810    & 427.        & 1           \\
            \midrule
        \end{tabular}
    \end{adjustbox}
    \caption{
        Performance and resource consumption of the HLF JSC models on the Xilinx UltraScale+ FPGAs with speedgrade \texttt{-2}. The models marked with $*$ only synthesized to netlist but did not perform place and route, and the \fmax and latency shown are based on the HLS target clock period. The models marked with $**$ did not report \fmax, but reported no timing violations at the specified target clock period. The results of DWN$^{***}$ are cited from~\cite{neuralut-assemble} instead of the original paper, as the original work omits preprocessing resource consumption. The design marked with $\dagger$ is obtained with the same model as the 9-cycle one in the row above it, but uses a less aggressive piplining strategy to achieve lower latency.
    }
    \label{tab:hlf_jsc}
    \vspace{-0.5cm}
\end{table*}

%% file: tables/plf_jsc.tex
\begin{table*}[htbp]
    \begin{adjustbox}{width=0.9\textwidth,center=\textwidth}
        \setlength\extrarowheight{-0.8pt}
        \begin{tabular}{lc|ccccccc}
            \multicolumn{8}{c}{\textbf{PLF JSC (3 features)}}                                                                                                                   \\
            \midrule
            Implementation                                      & Particles & Accuracy $\uparrow$ & Latency [cycles] & LUT       & DSP    & FF      & \fmax [MHz] & II [cycles] \\
            \midrule

            \textbf{HGQ (GNN)}                                  & 32        & 79.2\%              & 24 (108.8 ns)    & 159,238   & 0      & 91,027  & 220.5       & 1           \\
            \textbf{HGQ (GNN)}                                  & 32        & 78.5\%              & 20 (72.3 ns)     & 80,618    & 0      & 42,101  & 276.5       & 1           \\
            \texttt{QKeras, DS} [MLST'24]~\cite{ds-fpga}$^{*}$  & 32        & $\le 75.9$\%        & 26 (130 ns)      & 903,284   & 434    & 358,754 & $\sim 200$  & 2           \\
            \texttt{QKeras, GNN} [MLST'24]~\cite{ds-fpga}$^{*}$ & 32        & $\le 75.8$\%        & 32 (160 ns)      & 1,162,104 & 2,120  & 761,061 & $\sim 200$  & 3           \\
            \midrule
            \multicolumn{8}{c}{\textbf{PLF JSC (16 features)}}                                                                                                                  \\
            \midrule
            Implementation                                      & Particles & Accuracy $\uparrow$ & Latency [cycles] & (A)LUT    & DSP    & FF      & \fmax [MHz] & II [cycles] \\
            \midrule
            \textbf{HGQ (GNN)}                                  & 64        & 82.4\%              & 26 (122.5 ns)    & 244,515   & 0      & 112,993 & 212.2       & 1           \\
            \textbf{HGQ (GNN)}                                  & 64        & 80.7\%              & 9 (45.0 ns)      & 53,546    & 0      & 13,629  & 199.9       & 1           \\
            \textbf{HGQ (GNN)}                                  & 32        & 81.5\%              & 26 (119.8 ns)    & 238,255   & 0      & 116,039 & 217.1       & 1           \\
            \textbf{HGQ (GNN)}                                  & 32        & 80.2\%              & 9 (45.5 ns)      & 48,343    & 0      & 10,012  & 197.7       & 1           \\
            \textbf{HGQ (MLPM, Altera Agilex 7)}                & 64        & 81.1\%              & 10 (51.6 ns)     & 95,650    & 0      & 20,112  & 193.7       & 1           \\
            GNN U4 (TECS'24)~\cite{llgnn}$^{*}$                 & 50        & 80.9\%              & 130 (650 ns)     & 855k      & 8,945  & 201k    & $\sim 200$  & 100         \\
            GNN U5 (TECS'24)~\cite{llgnn}$^{*}$                 & 50        & 81.2\%              & 181 (905 ns)     & 815k      & 8,986  & 189k    & $\sim 200$  & 150         \\
            GNN J4 (TECS'24)~\cite{llgnn}$^{*}$                 & 30        & 78.4\%              & 58 (290 ns)      & 865k      & 8,776  & 138k    & $\sim 200$  & 30          \\
            GNN J5 (TECS'24)~\cite{llgnn}$^{*}$                 & 30        & 79.9\%              & 181 (905 ns)     & 911k      & 9,833  & 158k    & $\sim 200$  & 150         \\
            GNN (FPL'22)~\cite{que2022opt}$^{*}$                & 50        & 80.4\%              & 2132 (10660 ns)  & 1515k     & 12,284 & 533k    & $\sim 200$  & 650         \\
            GNN (FPL'22)~\cite{que2022opt}$^{*}$                & 30        & 78.7\%              & 382 (1910 ns)    & 1158k     & 11,504 & 246k    & $\sim 200$  & 400         \\
            \midrule
        \end{tabular}
    \end{adjustbox}
    \caption{
        The performance and resource consumption of the PLF JSC models on the Xilinx UltraScale+ FPGAs with speedgrade \texttt{-2} and Agilex 7 (part number \texttt{AGFB014R24A2E2V}) for the Quartus synthesized one. The models marked with $*$ only synthesized to netlist but did not perform place and route, and the \fmax and latency shown are based on the HLS target clock period.
    }
    \label{tab:plf_jsc}
    \vspace{-0.5cm}
\end{table*}

%% file: tables/tgc_svhn.tex
\begin{table*}[htbp]
    \begin{adjustbox}{width=0.9\textwidth,center=\textwidth}
        \setlength\extrarowheight{-0.8pt}
        \begin{tabular}{l|ccccccc}
            \multicolumn{8}{c}{\textbf{Muon tracking}}                                                                                                                 \\
            \midrule
            Implementation                                        & Resolution $\downarrow$ & Latency [cycles]  & LUT     & DSP   & FF     & \fmax [MHz] & II [cycles] \\
            \midrule
            \textbf{HGQ}                                          & 1.90 mrad               & 8 (47.4 ns)       & 41,830  & 0     & 10061  & 168.9       & 1           \\
            \textbf{HGQ}                                          & 2.03 mrad               & 6 (35.2 ns)       & 25,716  & 0     & 3455   & 170.3       & 1           \\
            \textbf{HGQ}                                          & 2.38 mrad               & 5 (28.7 ns)       & 14,789  & 0     & 3091   & 174.1       & 1           \\
            \texttt{QKeras} [NIMA'23]~\cite{tgc}$^{**}$           & 1.95 mrad               & 17 (106.3 ns)     & 37,867  & 1,762 & 8,443  & $>160$      & 1           \\
            \texttt{QKeras} [NIMA'23]~\cite{tgc}$^{**}$           & 2.04 mrad               & 13 (81.3 ns)      & 54,638  & 324   & 6,525  & $>160$      & 1           \\
            \texttt{QKeras} [NIMA'23]~\cite{tgc}$^{**}$           & 2.45 mrad               & 10 (62.5 ns)      & 28,526  & 24    & 2,954  & $>160$      & 1           \\
            \midrule
            \multicolumn{8}{c}{\textbf{SVHN classification}}                                                                                                           \\
            \midrule
            Implementation                                        & Accuracy $\uparrow$     & Latency [cycles]  & LUT     & DSP   & FF     & \fmax [MHz] & II [cycles] \\
            \midrule
            {\textbf{HGQ (hls4ml)}}                               & 93.8\%                  & 1,048 (5319.6 ns) & 66,056  & 52    & 24,207 & 197.0       & 1,029       \\

            {\textbf{HGQ (hls4ml)}}                               & 92.0\%                  & 1,060 (5272.4 ns) & 41,733  & 20    & 18,774 & 201.0       & 1,030       \\
            \texttt{QKeras} [MLST'21]~\cite{fast_cnn}$^{*}$       & 94.\%                   & 1,035 (5,175 ns)  & 111,152 & 174   & 32,554 & $\sim 200$  & 1,030       \\
            \texttt{QKeras} [MLST'21]~\cite{fast_cnn}$^{*}$       & 88.\%                   & 1,059 (5,295 ns)  & 38,795  & 70    & 14,802 & $\sim 200$  & 1,029       \\
            \texttt{DSP-Prune} [ICFPT'23]~\cite{dsp-prune}$^{**}$ & 92.4\%                  & 5,447 (43,576 ns) & 59,279  & 1,215 & 46,584 & $>125$      & N/A         \\
            \midrule
        \end{tabular}
    \end{adjustbox}
    \caption{
        The performance and resource consumption of the Muon tracking and SVHN classification models on the Xilinx UltraScale+ FPGAs with speedgrade \texttt{-2}. The models marked with $*$ only synthesized to netlist but did not perform place and route, and the \fmax and latency shown are based on the HLS target clock period. The models marked with $**$ did not report \fmax, but reported no timing violations at the specified target clock period.
    }
    \label{tab:tgc_svhn}
    \vspace{-0.5cm}
\end{table*}

%% file: tables/da_v_hls4ml.tex
\begin{table}
    \begin{adjustbox}{width=0.45\textwidth}
        \setlength\extrarowheight{-0.8pt}
        \begin{tabular}{lc|ccccc}
            \multicolumn{7}{c}{\textbf{HLF JSC OpenML}}                                                                       \\
            \midrule

            Implementation                                        & Accuracy & Latency  & LUT    & DSP & FF     & \fmax [MHz] \\
            \midrule
            \textbf{HGQ (hls4ml)}                                 & 76.9\%   & 53.0 ns  & 11,444 & 55  & 16,299 & 774.0       \\
            \textbf{HGQ (hls4ml)}                                 & 76.6\%   & 42.1 ns  & 7,101  & 27  & 9,858  & 831.3       \\
            \textbf{HGQ (hls4ml)}                                 & 75.6\%   & 30.5 ns  & 2,990  & 4   & 4,385  & 917.4       \\
            \textbf{HGQ (hls4ml)}                                 & 74.9\%   & 25.5 ns  & 1,933  & 0   & 2,859  & 939.8       \\
            \textbf{HGQ (hls4ml)}                                 & 73.2\%   & 19.2 ns  & 556    & 0   & 746    & 939.8       \\
            \texttt{DSP-Prune} [ICFPT'23]~\cite{dsp-prune}$^{**}$ & 76.3\%   & 105.0 ns & 5,504  & 175 & 3,036  & $>142.9$    \\
            \texttt{QKeras} [CoRR'21]~\cite{hls4ml}$^{*}$         & 76.2\%   & 45 ns    & 63,251 & 38, & 4,394  & $\sim 200$  \\
            \texttt{MetaML-Pro} [TRETS'26]~\cite{metamlpro}$^{*}$ & 76.1\%   & 50 ns    & 13,042 & 70  & N/A    & $\sim200$   \\
            \midrule
            \multicolumn{7}{c}{\textbf{HLF JSC CERNBox}}                                                                      \\
            \midrule
            Implementation                                        & Accuracy & Latency  & LUT    & DSP & FF     & \fmax [MHz] \\
            \midrule
            \textbf{HGQ (hls4ml)}                                 & 75.3\%   & 55.1 ns  & 12,377 & 56  & 17,654 & 743.5       \\
            \textbf{HGQ (hls4ml)}                                 & 75.1\%   & 50.0 ns  & 7,119  & 28  & 10,053 & 759.3       \\
            \textbf{HGQ (hls4ml)}                                 & 74.5\%   & 34.7 ns  & 4,141  & 3   & 5,716  & 864.3       \\
            \textbf{HGQ (hls4ml)}                                 & 73.2\%   & 23.4 ns  & 1,398  & 1   & 1,965  & 939.8       \\
            \textbf{HGQ (hls4ml)}                                 & 72.4\%   & 20.2 ns  & 1,045  & 0   & 1,445  & 939.8       \\
            \texttt{QKeras} [NMI'21]~\cite{qkeras}$^{*}$          & 74.8\%   & 55 ns    & 39,782 & 124 & 8,128  & $\sim 200$  \\
            \texttt{QKeras} [NMI'21]~\cite{qkeras}$^{*}$          & 72.3\%   & 55 ns    & 9,149  & 66  & 1,781  & $\sim 200$  \\
            \bottomrule
        \end{tabular}
    \end{adjustbox}
    \caption{Performance and resource consumption of the HLF JSC models synthesized with \texttt{hls4ml} compared to prior works using the same deployment backend. The models marked with $*$ only synthesized to netlist but did not perform place and route, and the \fmax and latency shown are based on the HLS target clock period. The models marked with $**$ did not report \fmax, but reported no timing violations at the specified target clock period.}
    \label{tab:da_v_hls4ml}
    \vspace{-0.9cm}
\end{table}